\documentclass{Interspeech}

\usepackage{subfigure}
\usepackage{amsthm,amsmath,amssymb}
\usepackage{mathrsfs}
\usepackage{multirow}
\usepackage{makecell}
\usepackage{enumitem}
\usepackage{listings}
\usepackage{color}
\usepackage[ruled, vlined, linesnumbered]{algorithm2e}

\interspeechcameraready

\title{MFLA: Monotonic Finite Look-ahead Attention for Streaming Speech Recognition}

\author[affiliation={1}]{Yinfeng}{Xia}
\author[affiliation={1}]{Huiyan}{Li}
\author[affiliation={2}]{Chenyang}{Le}
\author[affiliation={1}]{Manhong}{Wang}
\author[affiliation={1}]{Yutao}{Sun}
\author[affiliation={1}]{Xingyang}{Ma}
\author[affiliation={2}]{Yanmin}{Qian$^{\dagger}$}

\affiliation{}{}{$^1$Honor Device Co, Ltd, China}
\affiliation{}{}{$^2$Auditory Cognition and Computational Acoustics Lab}
\affiliation{}{}{MoE Key Lab of Artificial Intelligence, AI Institute}
\affiliation{}{}{School of Computer Science, Shanghai Jiao Tong University, Shanghai, China}
\email{\{xiayinfeng,lihuiyan\}@honor.com, \{yanminqian\}@sjtu.edu.cn}


\usepackage{comment}

\begin{document}

\maketitle

\begin{abstract}
% The application of large pre-trained speech models, such as Whisper, has shown significant potential in reducing training costs across various speech-related tasks. However, integrating these models into streaming systems remains a challenge. In this paper, we propose a novel prefix-to-prefix training framework to achieve streaming recognition by fine-tuning the pre-trained Whisper model. In particular, we introduce the Continuous Integrate-and-Fire mechanism to establish a quasi-monotonic alignment between continuous speech sequences and discrete text tokens. Subsequently, we design the Monotonic Finite Look-ahead Attention that allows each token to simultaneously attend to both the infinite left-context and finite right-context from the source speech sequences. Moreover, we adopt the widely-used wait-\textit{k} decoding strategy, which avoids introducing a complex decoding pipeline while maintaining consistency between training and testing conditions. Theoretical analysis and experimental results demonstrate that our approach achieves a controllable trade-off between latency and quality, making it suitable for diverse streaming applications. 

Applying large pre-trained speech models like Whisper has shown promise in reducing training costs for various speech tasks. However, integrating these models into streaming systems remains a challenge. This paper presents a novel prefix-to-prefix training framework for streaming recognition by fine-tuning the Whisper. We introduce the Continuous Integrate-and-Fire mechanism to establish a quasi-monotonic alignment between continuous speech sequences and discrete text tokens. Additionally, we design Monotonic Finite Look-ahead Attention, allowing each token to attend to infinite left-context and finite right-context from the speech sequences. We also employ the wait-\textit{k} decoding strategy to simplify the decoding process while ensuring consistency between training and testing. Our theoretical analysis and experiments demonstrate that this approach achieves a controllable trade-off between latency and quality, making it suitable for various streaming applications.

\end{abstract}

\section{Introduction}

As a framework for weakly supervised pre-training on large-scale datasets, Whisper~\cite{radford2023robust} has shown strong performance in multilingual recognition, but reveals a significant inference delay. Although methods such as knowledge distillation~\cite{gandhi2023distil,shao2023whisper} and speculative decoding~\cite{segal2024whisper}, have been proposed to improve inference speed, they do not alter the fundamental nature of the system as a sequence-to-sequence model, limiting its applicability only to offline systems. In contrast, online~(streaming) recognition systems employing prefix-to-prefix models are capable of satisfying latency requirements in certain specific scenarios, such as real-time subtitles. However, integrating the Whisper model into streaming systems presents significant challenges, primarily due to asynchronous processing problem~\cite{tian2020synchronous} and unreliable boundary transcription. 

% Specifically, the asynchronous processing challenge refers to the fact that target sequence prediction relies on the entire source sequence during the model training process, whereas online recognition must make predictions based on incomplete source prefixes, which can be alleviated by adopting monotonic attention. 

The challenge of asynchronous processing arises from the fundamental discrepancy between training and inference conditions: conventional sequence-to-sequence models leverage the full source context during training, while online systems must generate predictions incrementally based on partial source inputs. This inherent mismatch can be effectively addressed through monotonic attention mechanisms. Motivated by the observation of a roughly monotonic alignment between inputs and outputs, Raffel~\textit{et al.}~\cite{raffel2017online} proposed a differentiable approach that enables end-to-end alignment learning during training and linear-time decoding through hard monotonic constraints. Subsequent advancements introduced three principal variants: Monotonic Infinite Lookback Attention~(MILkA)~\cite{arivazhagan2019monotonic}, Monotonic Chunkwise Attention~(MoChA)~\cite{chiu2017monotonic}, and Monotonic Multihead Attention~(MMA)~\cite{ma2019monotonic}. 

Online recognition systems typically process input as fixed-length speech chunks, which can lead to unreliable transcription at chunk endpoints due to random truncation. To address these boundary-induced errors, the improved wait-\textit{k} policy~\cite{ma2018stacl} delays processing until the first \textit{k} chunks have been received and then output at a fixed rate \textit{r}; Macháček~\textit{et al.} integrated the Local Agreement policy~\cite{liu2020low} with self-adaptive latency into Whisper, identifying the longest common prefix between two consecutive chunks as stable hypotheses; Simul-Whisper~\cite{wang2024simul} halted decoding at the appropriate time and discarded unreliable transcriptions, this dual-protection strategy further minimized the risk of performance degradation in online systems.
%by considering the agreeing prefixes
%in online systems.
%Mach{\'a}{\v{c}}ek~\textit{et al.}

Although previous studies have proposed various solutions, these approaches often remain one-dimensional and struggle to balance latency and quality in streaming speech recognition systems. A key challenge in predicting the current token is its heavy influence from boundary ambiguity and acoustic similarity, which makes it inherently dependent on the appropriate context of the speech sequence. This dependency frequently leads to a predictable degradation in recognition quality when using conventional monotonic attention mechanisms. Furthermore, the improved wait-\textit{k} policy required setting a fixed output rate \textit{r} prior to decoding, making recognition latency and quality sensitive to variations in speaking speed and presence of silence segments, respectively. The Local Agreement policy introduced a higher fixed delay, while Simul-Whisper was hindered by a complex online decoding pipeline. These limitations highlight the need for more robust and flexible solutions for streaming speech recognition systems.

In this paper, we propose a novel prefix-to-prefix fine-tuning approach based on the pre-trained Whisper model, resulting in the fine-tuned model named Streaming-Whisper. The key contributions of this paper are summarized as follows:

\begin{enumerate}

\item We introduce a predictor based on the Continuous Integrate-and-Fire~(CIF) mechanism to estimate the number of target tokens, thereby establishing a quasi-monotonic alignment between continuous speech sequences and discrete tokens. 

\item We develop Monotonic Finite Look-ahead Attention~(MFLA) to enable each token to dynamically attend to both the infinite left-context and the finite right-context windows, which transforms the training paradigm from conventional sequence-to-sequence to a more efficient prefix-to-prefix framework.

% We develop Monotonic Finite Look-ahead Attention~(MFLA) to enable each token to attend to both the infinite left-context and the finite right-context, thereby transitioning from a conventional sequence-to-sequence framework to a more efficient prefix alignment approach.

\item We adopt the efficient wait-\textit{k} decoding strategy, which not only eliminates the complexity associated with additional decoding processes but also achieves a superior trade-off between latency and recognition quality compared to the state-of-the-art Local Agreement policy.

\end{enumerate}

\section{Methods}

\subsection{Overview}
% the decoder leverages the hidden states to generate the output text sequence $Y=\{y_{1},y_{2},...,y_{N}\}$ in an autoregressive manner,

\begin{figure} %[t]
  \centering
  \includegraphics[width=3.0in]{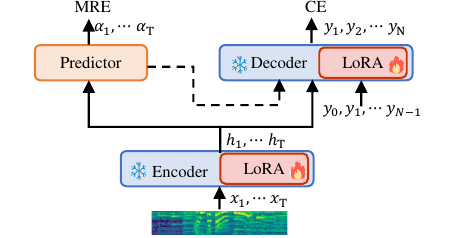}
  \caption{Structure of the proposed Streaming-Whisper.}
  \label{Online-Whisper}
\end{figure}

As shown in Figure~\ref{Online-Whisper}, the proposed Streaming-Whisper includes three modules: encoder, decoder, and predictor. The encoder converts the input speech sequence $X=\{x_{1},x_{2},...,x_{T}\}$ into a hidden state sequence $H=\{h_{1},h_{2},...,h_{T}\}$,  and defines $h_{1:T} = \mathit{f}(x_{1:T})$; the decoder employs the hidden states to produce the output sequence $Y=\{y_{1},y_{2},...,y_{N}\}$ through an autoregressive process, and defines $y_{i}= \mathit{g}(y_{i-1},h_{1:T})$.  The predictor is adopted to establish the number of target tokens and guide the generation of MFLA, which will be discussed in detail in Section~\ref{predictor} and Section~\ref{monotonic-attention}.

\subsection{Predictor}
\label{predictor}

The predictor consists of two linear layers and two ReLU activation layers, which can predict the token weight $\alpha_{1:T}$ of each hidden state $h_{1:T}$. The expression of the predictor function is defined as $\alpha_{j}=e(h_{j})$. We accumulate the weights $\alpha_{1:T}$ to determine the number of target tokens, and the loss function is defined as the Mean Relative Error~(MRE) loss. 

%$L_{MRE} = \left|\frac{N-\sum_{j=1}^{T}\alpha_{j}}{N}\right|$.

Subsequently, we introduce the CIF mechanism to establish a quasi-monotonic alignment between continuous speech signals and their corresponding discrete tokens, which delineates the temporal boundaries (left and right) for each target token. This mechanism provides three significant benefits: (1) during the training phase, it manages the finite right-context window to train online speech recognition systems by guiding the generation of MFLA; (2) during the incremental decoding stage, it allows for tracking the streaming decoding process and stopping it at the appropriate time to prevent unreliable boundary transcriptions;~(3) it can monitor the entire decoding trajectory, thus mitigating common decoding repetition problems~\cite{li2024transcription}.

% \begin{enumerate}[]
% \item During the training stage, it manages the finite right context to train online speech recognition systems, guiding the generation of MFLA.
% \item During the incremental decoding stage, it allows for tracking the streaming decoding process and stopping it at the appropriate time to prevent unreliable boundary transcriptions.
% \item It can monitor the entire decoding process, thus avoiding common decoding repetition problem~\cite{li2024transcription}.
% \end{enumerate}

According to~\cite{dong2020cif,gao2022paraformer}, the weight $\alpha$ is scaled by the target length $N$ during training, while weight $\alpha$ is used directly during inference.

\subsection{Monotonic Attention}
\label{monotonic-attention}

\begin{figure}%[t]
\centering
\subfigure[]{\label{Fig1_1a}
\begin{minipage}[t]{0.30\linewidth}
\centering
\includegraphics[width=0.90in]{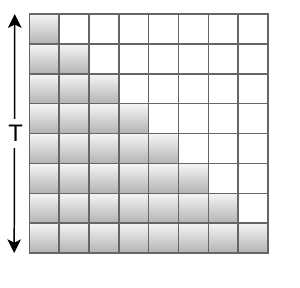}
\end{minipage}%
}%
\subfigure[]{\label{Fig1_1b}
\begin{minipage}[t]{0.30\linewidth}
\centering
\includegraphics[width=0.90in]{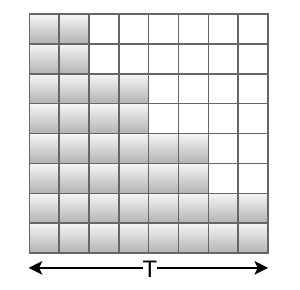}
\end{minipage}%
}%
\subfigure[]{\label{Fig1_1c}
\begin{minipage}[t]{0.30\linewidth}
\centering
\includegraphics[width=0.90in]{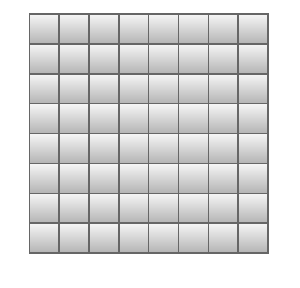}
%\caption{fig2}
\end{minipage}
}%
\quad
\subfigure[]{\label{Fig1_1d}
\begin{minipage}[t]{0.30\linewidth}
\centering
\includegraphics[width=0.90in]{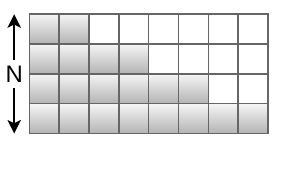}
\end{minipage}%
}%
\subfigure[]{\label{Fig1_1e}
\begin{minipage}[t]{0.30\linewidth}
\centering
\includegraphics[width=0.90in]{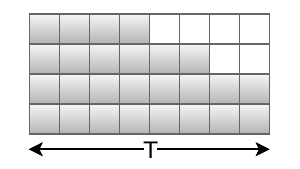}
\end{minipage}%
}%
\subfigure[]{\label{Fig1_1f}
\begin{minipage}[t]{0.30\linewidth}
\centering
\includegraphics[width=0.90in]{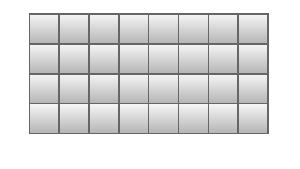}
%\caption{fig2}
\end{minipage}
}%
\centering
\caption{MoChA in encoder and MFLA in decoder.}
\label{mask}
\end{figure}

In an offline speech recognition system, the encoder processes the entire speech sequence as input and outputs a corresponding sequence of hidden states; consequently, its self-attention mechanism operates under full-attention during training. In contrast, online systems employ encoders that sequentially process a series of fixed-size speech chunks. To adapt to this setting, we replace the conventional convolutional layers in front of the Whisper's encoder transformer with causal convolution layers. Additionally, we implement MoChA~\cite{chiu2017monotonic}, which restricts the attention within each chunk to only the current chunk and previous chunks. Figures~\ref{Fig1_1a} and \ref{Fig1_1b} illustrate the MoChA with chunk sizes of 1 and 2, respectively. 

% To prevent future context leakage, the identical causal attention constraints also need to be implemented on the decoder module. Through the CIF mechanism, the hidden state sequence undergoes decomposition into $N$ segments, establishing monotonic frame-to-token alignment through consecutive frame mapping. While the monotonic infinite look-back attention mechanism facilitates so-called \textit{"real-time"} recognition, two critical challenges must be addressed: (1) the inherently ambiguous between speech segments corresponding to individual tokens, and (2) acoustic similarities between certain tokens. These lead to obvious context-dependency in speech recognition.

To preserve strict causality, we also implement consistent causal attention constraints in the decoder module to prevent exploiting excessively long future context. The CIF mechanism dynamically segments the hidden state sequence into $N$ segments through frame aggregation, ensuring monotonic alignment between acoustic frames and text tokens. While the MILkA mechanism facilitates so-called \textit{"real-time"} recognition, two critical challenges emerge: (1) the inherently ambiguous between acoustic segments and their corresponding individual tokens, and (2) phonetic confusability among acoustically similar tokens. These phenomena induce increased context sensitivity that significantly impacts recognition robustness.

% This mechanism enables each token to interact with both infinite left context and constrained right context during training, thus establishing controlled look-ahead capacity.

Drawing inspiration from the wait-\textit{k} policy employed in simultaneous interpretation, which maintains a consistent \textit{k} words output lag relative to input, we propose a finite-delay monotonic attention mechanism. Through a finite look-ahead window, the mechanism enables controlled access to finite right-context during training while maintaining unbounded left-context accessibility. We term this mechanism Monotonic Finite Look-ahead Attention~(MFLA), here we assume that each token corresponds to only $2$ frames in the hidden state sequence, Figures~\ref{Fig1_1d} and \ref{Fig1_1e} depict MFLA mechanisms with the look-ahead window span of 1 and 2, respectively. 

% This mechanism enables controlled right-context utilization through constrained look-ahead windows while preserving infinite left-context access during training.

% This mechanism enables controlled right-context utilization through constrained look-ahead windows while preserving infinite left-context access during training. 

Indeed, by examining the look-ahead implementation of attention during both training and testing, we can regard an offline system as a specific case of an online system. That is, the chunk size in MoChA and look-ahead span in MFLA are both $\infty$, as shown in Figures~\ref{Fig1_1c} and \ref{Fig1_1f}, respectively. 

\subsection{Online Decoding Method}

We reformulate online decoding of Streaming-Whisper as a read-write policy problem: the system should initiate responses when it has gathered sufficient information and cease when the current information is inadequate. As discussed in Section~\ref{predictor} and \ref{monotonic-attention}, we can map continuous speech sequences to discrete tokens and adopt a look-ahead strategy to focus on finite right-context, based on the CIF mechanism and MFLA, respectively. This approach enables our online decoding method to mimic simultaneous interpretation, which not only facilitates the direct application of the wait-\textit{k} decoding policy but also ensures the consistency of the training method and the inference process. The online decoding method is detailed in Algorithm 1.

\begin{algorithm}
  % \label{algorithm.1}
  \SetKwData{Left}{left}\SetKwData{This}{this}\SetKwData{Up}{up}
  \SetKwFunction{Union}{Union}\SetKwFunction{FindCompress}{FindCompress}
  \SetKwInOut{Input}{input}\SetKwInOut{Output}{output}

  \Input{Speech sequence $ X =\{x_{1},x_{2},...,x_{T}\}$, $\textit{k}$}
  \Output{Text token}  %$Y=\{y_{1},y_{2},...,y_{N}\}$
  \BlankLine
  Initialize \emph{$j \leftarrow 1 , i \leftarrow 0, \alpha \leftarrow 0, y_{0} \leftarrow \langle \textit{sos} \rangle $} \\  
  %\emph{special treatment of the first line}\;
  \While{$j \leq T$}{
    \emph{$h_{j} \leftarrow \boldsymbol{f}(x_{j})$} \quad $\blacktriangleright$ \textbf{\textit{Read action}}\\
    \emph{$\alpha \leftarrow \alpha + \boldsymbol{e}(h_{j})$} \\
    \While{$\alpha > \textit{k} $}{
      \emph{$y_{i+1} \leftarrow \boldsymbol{g}(y_{i},h_{1:j})$} 
      \quad $\blacktriangleright$   \textbf{\textit{Write action}} \\
      \emph{$i \leftarrow i+1 $} \\
      \emph{$\alpha \leftarrow \alpha-1 $} \\
      }
    \emph{$j \leftarrow j+1$}  \\
  } 
  \While {$y_{i} \neq \langle \textit{eos} \rangle $}{
    \emph{$y_{i+1} \leftarrow \boldsymbol{g}(y_{i},h_{1:T})$} \quad $\blacktriangleright$  \textbf{\textit{Write action}} \\
    \emph{$i \leftarrow i+1$}
  }
  \Return {$y_{1:i}$}
  \caption{online decoding method}
  \label{sdm}
\end{algorithm}

\section{Experimental Setup}

% \subsection{Model}
% Our experimental framework encompasses Whisper architectures at multiple scales, including Small, Medium, and Large-V3-Turbo. Among these configurations, Whisper-Large-V3-Turbo has only four decoding layers, which can achieve fast autoregressive decoding and is very suitable for streaming recognition scenarios.

% Additionally, we extend our approach to SpeechLLM to enhance streaming recognition performance. Inspired by BESTOW~\cite{chen2024bestow}, the speech encoder is initialized by the Whisper-Large-V3 encoder, the LLM is initialized by the Qwen2.5-3B-Instruct~\cite{yang2024qwen2} model, and the adapter layer consists of two layers of trainable transformer-like self-attention and cross-attention blocks. 

\subsection{Data}

Our training and evaluation data are constructed from various open-source datasets, including WenetSpeech4TTS~\cite{ma2024wenetspeech4tts} (where a portion of each subset is reserved for the testset), LibriSpeech~\cite{panayotov2015librispeech}, Multilingual Librispeech~(MLS)~\cite{pratap2020mls}, and VoxPopuli~\cite{wang2021voxpopuli}, covering four languages: Chinese~(cn), English~(en), German~(de) and Spanish~(es). In particular, for WenetSpeech4TTS, we only use the Premium subset during model training to ensure balanced data distribution.

% The proposed approach is evaluated on WenetSpeech4TTS, LibriSpeech and MLS datasets.

% as well as larger datasets like WenetSpeech4TTS.Premium~\cite{ma2024wenetspeech4tts} (partly used as the testset) and Emilia~\cite{emilia},  and covers four languages: Chinese~(cn), English~(en), German~(de) and Spanish~(es). For Streaming-Whisper, we use smaller datasets and WenetSpeech4TTS.Premium; for SpeechLLM, we only use Chinese and English parts from the full dataset. The proposed method is evaluated on WenetSpeech4TTS, LibriSpeech and MLS datasets.

\subsection{Training Setting}

The predictor is randomly initialized, and we employ Low-Rank Adaptation~(LoRA)~\cite{hu2021lora} to freeze the parameters of both the speech encoder and the decoder throughout the fine-tuning process. Inspired by WeNet~\cite{yao2021wenet}, the fine-tuning process of the model employs a hybrid-attention mechanism that combines full-attention and monotonic-attention. For MoChA, the chunk size follows a uniform distribution within the interval $[32, 128]$; while for MFLA, the look-ahead span is modeled as a Poisson distribution with $\lambda=3$. Considering the dependence of the MFLA generation process on the predictor, we adopt a two-stage fine-tuning strategy. In the first stage, we only use full-attention to train the decoder; in the second stage, we translate the decoder's attention mechanism from full-attention to hybrid-attention. The total loss comprises the MRE loss of the predictor and the Cross-Entropy~(CE) loss of the decoder, which is weighted by $\gamma=5$.

\subsection{Decoding and Evaluation}

% Since our model is built on hybrid-attention training, it supports both offline decoding and online incremental decoding modes. For incremental decoding, Liu~\cite{liu2020low} notes that the decoding of a new block can either continue from the previously buffered decoder state or commence after forcing decoding with the committed tokens. In the absence of specific instructions, we employ the forced decoding method with the committed tokens in streaming scenarios. In the following experiments, we perform offline decoding and online incremental decoding using the greedy decoding strategy, all within a single model framework.

Our model architecture, leveraging hybrid-attention training, inherently supports dual decoding paradigms: offline decoding and online incremental decoding. Regarding the incremental decoding approach, Liu~\cite{liu2020low} identifies two different forms: (1) initialization through forced decoding with committed tokens, and (2) continuation from the buffered decoder state. In our experimental setup, we implemented the wait-3 decoding policy with forced token commitment as our default online decoding strategy, and conducted both offline and online decoding within a unified model framework by utilizing greedy search strategy.

% In our experimental setup, we employ greedy search strategy to perform offline and online decoding within a unified model framework, where online decoding encompasses the above two forms.

%In our experimental setup, we implement the wait-3 policy with forced token commitment as our default online decoding strategy without explicit instruction, and conduct both offline and online decoding within a unified model framework by utilizing greedy search strategy.

 % In our experimental setup, we employ the greedy search strategy to perform offline or online decoding within a unified model framework, while the online decoding uses these two incremental decoding forms to compare the differences.

 % implement the wait-3 policy with forced token commitment as our default online decoding strategy without explicit instruction, and conduct both offline and online decoding within a unified model framework by utilizing greedy search strategy.

% In our experimental setup, we employ the wait-3 policy combined with forced token commitment as the default configuration for online decoding in scenarios without explicit instructions, while leveraging greedy search as the unified decoding strategy to enable both offline and online processing within a single architectural framework.

% conduct both offline and online decoding within a unified model framework by utilizing greedy search strategy.

The generated transcriptions and ground truth labels are normalized using Whisper's open-source text normalizer, and the Python library edit distance~\cite{yujian2007normalized} is used to evaluate the word error rates (WER). All inference stages are performed on a single NVIDIA L20 GPU with 48 GB of memory.  

% H800 GPU with 80GB of memory. 
% L20 GPU with 48GB of memory.

\section{Experimental Results}

\subsection{Architecture Experiment}
We implement the Streaming-Whisper framework in various scale architectures, including Small, Medium, Large-V3 and Large-V3-Turbo. Table~\ref{t1} presents the WERs(\%) of different decoding methods in different models. The experimental results indicate that online decoding method exhibits consistent performance degradation compared to offline decoding with respective performance degradation of 1.72\%, 1.56\%, 1.18\%, and 1.54\% for the respective models of corresponding scales.

%We implemente the Streaming-Whisper framework across various scale architectures, including Small, Medium, Large-V3, and Large-V3-Turbo. Table~\ref{t1} presents the WERs(\%) of different decoding methods in different models. The experimental results demonstrate that online decoding method exhibit predictable performance degradation compared to their offline counterparts, with respective performance degradation of 1.72\%, 1.56\%, 1.18\%, and 1.54\% on the models of corresponding scales, respectively.
%Specifically, the performance degradation observed is 1.72%, 1.56%, 1.18%, and 1.54% for the respective models of corresponding scales.

% Among these frameworks, the Whisper-Large-V3-Turbo model features only four decoding layers, which makes it highly suitable for streaming recognition tasks. Therefore, we conduct ablation experiments under this framework to compare different decoding strategies, see Sections ~\ref{result2} and ~\ref{result3} for details.
% , including local-agreement policy, wait pilcies with different \textit{k},
% Therefore, we conduct ablation experiments on different decoding strategies within this framework, detailed in Section~\ref{result2} and~\ref{result3}. 

\subsection{Ablation Experiment}

Among the evaluated architectures, the Whisper-Large-V3-Turbo model incorporates merely four decoding layers, rendering it particularly suitable for streaming recognition scenarios. Therefore, we conduct ablation experiments within this framework to compare the accuracy, latency, and computational complexity of different online decoding methods. The results are shown in Table~\ref{t2} with the Local Agreement policy established as our baseline.

% Among these frameworks, the Whisper-Large-V3-Turbo model features only four decoding layers, which makes it highly suitable for streaming recognition tasks. Therefore, we conduct ablation experiments under this framework to compare the accuracy, latency, and computational complexity of different online decoding methods. We regard the Local Agreement policy as the baseline, and the results are shown in Table~\ref{t2}.

\subsubsection{Accuracy}
% Table~\ref{t2} shows the average WER across all datasets for various decoding strategies. 
%We use the average WER as an indicator to evaluate the recognition accuracy.  

Under the wait-\textit{k} policy, the WER demonstrates a monotonic decrease with increasing \textit{k} values, implying that expanding the right-context window enhances recognition accuracy at the cost of increased latency. It is particularly noteworthy that the Local Agreement policy surpasses the wait-\textit{k} approach in performance, primarily because it incorporates an implicit error correction mechanism through consensus-building across consecutive speech segments, thereby substantially enhancing hypothesis reliability. In addition, the 1.18\% performance gap between the wait-$\infty$ and offline decoding method reveals that the LoRA-based fine-tuning approach demonstrates limited effectiveness in enhancing the encoder's processing of streaming speech.

% Notably, the Local Agreement policy outperforms wait-\textit{k} policy because it implements an implicit error correction paradigm by establishing consensus between consecutive speech chunks, which can improve the hypothesis reliability.

%but DAL has relative performance reduction of 80\%, 70\%, and 50\%, respectively.

%across various Whisper architectures, with a chunk length of 1 second.
\begin{table*}[t]
\centering
\caption{The WERs(\%) of offline and online decoding methods on testsets, with a chunk length of 1 second for online decoding.} 
% The wait-3 policy is adopt as the online decoding method, where $\ddagger$ and $\ddagger$ represent incremental decoding strategies based on forced token commitment and buffered state continuation, respectively.
\label{t1}
% \begin{tabular}{cccclcclccllllc}
\begin{tabular}{ccccccccccccccc}
\hline
\multirow{2}{*}{Architecture}  & \multirow{2}{*}{Methods} & \multicolumn{2}{c}{WenetSpeech4TTS} &  & \multicolumn{2}{c}{Librispeech.test} &  & \multicolumn{2}{c}{MLS} &  & \multicolumn{3}{c}{VoxPopuli} & \multicolumn{1}{l}{\multirow{2}{*}{\textit{\textbf{Avg}}}} \\ \cline{3-4} \cline{6-7} \cline{9-10} \cline{12-14}
                                &                          & Premium          & Standard         &  & clean             & other            &  & de         & es         &  & en       & de       & es      & \multicolumn{1}{l}{}                                       \\ \hline
\multirow{2}{*}{Small}          

& Offline                  & 5.25             & 6.57             &  & 4.39              & 8.20             &  & 7.67       & 4.99       &  &8.08          &14.48          &9.32         & 7.66   \\
& Online                    & 6.47             & 8.05             &  & 4.86              & 10.60             &  & 9.67       & 6.50       &  &9.32          &17.72          &11.27         & 9.38  \\ 
% & Wait-3 $\dagger$          & 6.60             & 8.17             &  & 5.59              & 11.25             &  & 12.17       & 8.41       &  &9.85          &17.79          &12.05         & 10.21  \\  
\hline
\multirow{2}{*}{Medium}         

& Offline                  & 3.91             & 5.28             &  & 4.09              & 6.80             &  & 5.29       & 3.35       &  &7.03          &11.22          &7.60         & 6.06   \\
& Online                   & 5.45             & 6.86             &  & 4.41              & 8.79             &  & 6.97       & 4.55       &  &8.28          &13.77          &9.58         & 7.62  \\ 
% & Wait-3$\dagger$         & 4.33             & 5.47             &  & 3.40              & 7.83             &  & 8.38       & 5.48       &  &8.14          &15.72          &9.95         & 7.63  \\ 
\hline
\multirow{2}{*}{Large-V3} 
& Offline                  & 3.47             & 4.67             &  & 3.86             & 5.73             &  & 4.53       & 2.86       &  &6.88          &10.66          &7.11         & 5.53   \\
& Online                   & 4.54             & 6.05             &  & 3.97              & 7.20            &  & 5.86       & 3.66       &  &7.89          &12.57          &8.62         & 6.71  \\ 
% & Wait-3$\dagger$         & 4.77             & 6.30             &  & 4.25              & 7.52             &  & 6.36       & 4.26       &  &7.86          &12.93          &8.74         & 7.00  \\ 
\hline
\multirow{2}{*}{\makecell[c]{Large-V3-\\Turbo}} 
& Offline                  & 4.11             & 5.34             &  & 3.76              & 6.02             &  & 4.42       & 2.65       &  & 6.93     & 10.36    & 7.11    & 5.63                                                       \\
& Online                   & 5.47             & 7.21             &  & 4.23              & 8.19             &  & 6.11       & 3.67       &  &8.14          &12.73          & 8.77        & 7.17  \\
% & Wait-3$\dagger$         & 5.75             & 7.53            &  & 4.41              & 8.34             &  & 6.62       & 3.85       &  &8.21          &13.35          & 9.01        & 7.45    \\ 
\hline
\end{tabular}
\end{table*}

\subsubsection{Latency}
% \label{result2}
% In addition, we also evaluate the latency and floating point operations~(FLOPs) of various decoding methods. 
% Following~\cite{wang2024simul}, we adopt the Differentiable Average Lagging~(DAL) metric~\cite{cherry2019thinking} to assess system latency. 

We adopt the Differentiable Average Lagging~(DAL) metric~\cite{cherry2019thinking} indicator to evaluate response latency. DAL can quantify the average latency relative to a streaming system across all tokens. Assuming the input speech length is $N_{s}$ and the number of output tokens is $N_{t}$, the ideal streaming policy generates a token every $d = N_{s}/N_{t}$ seconds, the token $t$ is generated at time $g(t)$. The calculation method of DAL as follows:

\begin{equation}
g^{\prime}{ }_{d}(t)=\left\{\begin{array}{ll}
g(t) & t=1 \\
\max \left(g(t), g^{\prime}_{d}(t-1)+d\right) & t>1
\end{array}\right.
\end{equation}

\begin{equation}
\begin{aligned}
{\rm DAL} &=\frac{1}{N_{t}} \sum_{t=1}^{N_{t}} g^{\prime}{ }_{d}(t) - (t-1)d 
\end{aligned}
\end{equation}

In fact, we can theoretically derive the latency for the Local Agreement and wait-\textit{k} policies. In ideal computation-unaware scenarios, let $N_{c}$ denote the length of an input speech chunk. We present the derived DAL expressions for both policies in Equations~\ref{lg_dal} and \ref{waitk_dal}, respectively. The variables in the DAL expressions are defined as follows: $N_{c}$ represents the input chunk length preset by the system; $d$ denotes the speaking rate, which is not controlled by the system; and $k$ indicates the number of right-context windows, with a minimum value of 1. It is worth noting that the Local Agreement policy can only reduce latency by shortening the input chunk length, while the wait-\textit{k} policy provides better adaptability by adjusting the parameter $k$. Moreover, due to the continuity of speech features, the parameter $k$ is highly flexible and can even take decimals.

%, enabling a trade-off between latency and quality. 

% By selecting a smaller value of $k$, we can ensure that the latency of the wait-\textit{k} policy is lower than the Local Agreement. Moreover, due to the continuity of speech features, the value of $k$ is highly flexible and can even assume decimal values.

\begin{equation}\label{lg_dal}
\begin{aligned}
{\rm DAL}_{local-agreement} &= \frac{3}{2}N_{c} + \frac{d}{2}
\end{aligned}
\end{equation}

\begin{equation}\label{waitk_dal}
\begin{aligned}
{\rm DAL}_{wait-k} &= \frac{1}{2}N_{c} + (k-\frac{1}{2})d
\end{aligned}
\end{equation}

% Ideally, the minimum DAL of Local Agreement and wait-\textit{k} policies are $\frac{3}{2}N_{c}$ and $\frac{1}{2}N_{c}$ respectively.

In streaming scenarios, the operating conditions of encoders employing different decoding methods are identical, thus encoder latency is not considered in the computation-aware DAL. As shown in Table~\ref{t2}, compared to the Local Agreement policy, the WER of the wait-\textit{k} policy is degraded by 0.53\%, 0.19\%, and 0.11\% for \textit{k} is 1, 2, and 3, respectively. However, the relative delay is significantly reduced by 43.63\%, 29.09\%, and 14.54\% for the corresponding \textit{k} values. This demonstrate that our approach can effectively balance the real-time requirements and quality constraints in various online systems by adjusting the value of \textit{k}. 
% Moreover, due to the continuity of speech features, the value of $k$ is highly flexible and can even assume decimal values.

\subsubsection{Computational Complexity}
\label{result3}
% We also compared the FLOPs of the two online decoding strategies. As demonstrated in Table~\ref{t2}, the decoding operation of wait-\textit{k} is lower than that of the Local Agreement. In addition, we can adopt incremental decoding strategy with buffer state continuation to avoid redundant computation caused by frequent decoder buffer resets. Compared with wait-3 decoding policy, wait-3$\dagger$ can reduce 60\% of redundant computation in the decoder at the cost of only 0.14\% performance degradation.

We also performed a comparative analysis of the computational complexity, measured in FLOPs, for these decoding strategies in the decoder. As demonstrated in Table~\ref{t2}, the decoding operation of wait-\textit{k} exhibits lower computational overhead compared to the Local Agreement. Furthermore, we adopt the incremental decoding strategy with buffer state continuation to avoid redundant computation caused by frequent decoder buffer resets. Specifically, compared to the wait-3 policy, wait-3$\dagger$ achieves a 60.86\% reduction in redundant computation within the decoder at the cost of only 0.14\% performance degradation.

\begin{table}[t]
\centering
\caption{The average metrics of different online decoding methods in WER(\%), DAL(s) and FLOPs(G), $\dagger$ represents buffered state continuation incremental decoding strategy. The Local Agreement online incremental decoding method is considered as the Baseline.}
\label{t2}
\begin{tabular}{cccc}
\hline
Methods         & \multicolumn{1}{l}{WER} & \multicolumn{1}{l}{DAL} & \multicolumn{1}{l}{FLOPs} \\ \hline
% Offline         & 5.63                    & 6.71                    & 12.85                     \\
Baseline        & 7.06                    & 1.65                    & 37.56                     \\
Wait-1          & 7.59                    & 0.93                    & 34.35                     \\
Wait-2          & 7.25                    & 1.17                    & 33.48                     \\
Wait-3          & 7.17                    & 1.41                    & 32.63                     \\
Wait-3$\dagger$ & 7.31                    & 1.41                    & 12.77                     \\
Wait-5          & 7.10                    & 1.87                    & 31.06                     \\
Wait-$\infty$   & 6.81                    & 6.71                    & 12.85                     \\ \hline
\end{tabular}
\end{table}

\subsection{SpeechLLM}

We extend our approach to SpeechLLM to enhance streaming recognition performance. Inspired by BESTOW~\cite{chen2024bestow}, the speech encoder and LLM are initialized by the Whisper-Large-V3 and Qwen2.5-3B-Instruct~\cite{yang2024qwen2} model, respectively; the adapter layer consists of two layers of trainable transformer-like self-attention and cross-attention blocks. Compared to existing LLM-based speech streaming recognition systems~\cite{jia2024efficient,tsunoo2024decoder}, our approach eliminates both training data preprocessing requirements and complex decoding pipeline construction, resulting in a more streamlined and efficient framework. 

% We employ LoRA to freeze the parameters of speech encoder and LLM throughout the fine-tuning process. In addition, we construct larger training datasets, including WenetSpeech4TTS and Emilia~\cite{emilia}, focusing on Chinese and English.

As evidenced in Table~\ref{t3}, SpeechLLM demonstrates superior recognition performance compared to Whisper across both offline and online processing paradigms, underscoring the substantial potential of integrating speech recognition systems with LLMs. Compared with offline decoding, the performance of SpeechLLM's online decoding decreases by 0.98\%.

\begin{table}[t]
\centering
\caption{The WERs(\%) of different decoding methods on SpeechLLM, with the chunk length of 1 second.}
\label{t3}
\begin{tabular}{cccccc}
\hline
\multirow{2}{*}{Methods} & \multicolumn{2}{c}{WenetSpeech4TTS} & \multicolumn{2}{c}{Librispeech.test} & \multicolumn{1}{l}{\multirow{2}{*}{\textit{\textbf{Avg}}}} \\ \cline{2-5}
                         & Premium          & Standard          & clean          & other         & \multicolumn{1}{l}{}                                       \\ \hline
Offline                  & 2.77        & 3.72        & 1.92           & 4.15          & 3.14                                                      \\
% L.A.                     & 3.35        & 4.46        & 2.35           & 6.00          & 1.71                                                      \\
Online                & 3.41        & 4.51        & 2.38           & 6.19          & 4.12                                                      \\
% Online$\dagger$                & 3.66        & 4.82        & 2.61           & 6.59          & 4.42                                                      \\ 
\hline
\end{tabular}
\end{table}

\section{Conclusions and Discussions}

In this paper, we propose MFLA, an attention mechanism that enables each token to attend to both the infinite left-context and finite right-context in the speech sequence. This mechanism allows the training approach to shift from a sequence-to-sequence paradigm to a prefix-to-prefix paradigm, thereby facilitating real-time speech recognition through fine-tuning of the pre-trained Whisper model. Furthermore, we employ the fundamental wait-\textit{k} decoding policy to enable control of the latency-quality trade-off in streaming scenarios.

% Compared to existing large speech streaming recognition systems~\cite{liu2020low,jia2024efficient,tsunoo2024decoder}, our approach eliminates both training data preprocessing requirements and complex decoding pipeline construction, resulting in a more streamlined and efficient framework.

% However, our approach still encounters two challenges. Firstly, the network structure and loss constraints of the predictor are overly simplistic, resulting in irregular deviations in CIF predictions at the frame level. Secondly, the fine-tuning approach based on LoRA has demonstrated limited effectiveness in enhancing the performance of streaming encoders. In future work, we will explore methods to address these issues.

However, our approach still encounters primary limitations. First, the network structure and loss constraints of the predictor are overly simplistic, leading to biased estimation of frame-level token weights. Second, the LoRA-based fine-tuning method has demonstrated limited effectiveness in enhancing the encoder's processing of streaming speech. We will explore methods to address these issues in future work.

\section{Acknowledgements}

This work was supported in part by China NSFC projects under Grants 62122050 and 62071288, in part by Shanghai Municipal Science and Technology Commission Project under Grant 2021SHZDZX0102.

%\clearpage
\bibliographystyle{IEEEtran}
\bibliography{main}

\begin{thebibliography}{10}
\providecommand{\url}[1]{#1}
\csname url@samestyle\endcsname
\providecommand{\newblock}{\relax}
\providecommand{\bibinfo}[2]{#2}
\providecommand{\BIBentrySTDinterwordspacing}{\spaceskip=0pt\relax}
\providecommand{\BIBentryALTinterwordstretchfactor}{4}
\providecommand{\BIBentryALTinterwordspacing}{\spaceskip=\fontdimen2\font plus
\BIBentryALTinterwordstretchfactor\fontdimen3\font minus \fontdimen4\font\relax}
\providecommand{\BIBforeignlanguage}[2]{{%
\expandafter\ifx\csname l@#1\endcsname\relax
\typeout{** WARNING: IEEEtran.bst: No hyphenation pattern has been}%
\typeout{** loaded for the language `#1'. Using the pattern for}%
\typeout{** the default language instead.}%
\else
\language=\csname l@#1\endcsname
\fi
#2}}
\providecommand{\BIBdecl}{\relax}
\BIBdecl

\bibitem{radford2023robust}
A.~Radford, J.~W. Kim, T.~Xu, G.~Brockman, C.~McLeavey, and I.~Sutskever, ``Robust speech recognition via large-scale weak supervision,'' in \emph{International conference on machine learning}.\hskip 1em plus 0.5em minus 0.4em\relax PMLR, 2023, pp. 28\,492--28\,518.

\bibitem{gandhi2023distil}
S.~Gandhi, P.~von Platen, and A.~M. Rush, ``Distil-whisper: Robust knowledge distillation via large-scale pseudo labelling,'' \emph{arXiv preprint arXiv:2311.00430}, 2023.

\bibitem{shao2023whisper}
H.~Shao, W.~Wang, B.~Liu, X.~Gong, H.~Wang, and Y.~Qian, ``Whisper-kdq: A lightweight whisper via guided knowledge distillation and quantization for efficient asr,'' \emph{arXiv preprint arXiv:2305.10788}, 2023.

\bibitem{segal2024whisper}
Y.~Segal-Feldman, A.~Shamsian, A.~Navon, G.~Hetz, and J.~Keshet, ``Whisper in medusa's ear: Multi-head efficient decoding for transformer-based asr,'' \emph{arXiv preprint arXiv:2409.15869}, 2024.

\bibitem{tian2020synchronous}
Z.~Tian, J.~Yi, Y.~Bai, J.~Tao, S.~Zhang, and Z.~Wen, ``Synchronous transformers for end-to-end speech recognition,'' in \emph{ICASSP 2020-2020 IEEE International Conference on Acoustics, Speech and Signal Processing (ICASSP)}.\hskip 1em plus 0.5em minus 0.4em\relax IEEE, 2020, pp. 7884--7888.

\bibitem{raffel2017online}
C.~Raffel, M.-T. Luong, P.~J. Liu, R.~J. Weiss, and D.~Eck, ``Online and linear-time attention by enforcing monotonic alignments,'' in \emph{International conference on machine learning}.\hskip 1em plus 0.5em minus 0.4em\relax PMLR, 2017, pp. 2837--2846.

\bibitem{arivazhagan2019monotonic}
N.~Arivazhagan, C.~Cherry, W.~Macherey, C.-C. Chiu, S.~Yavuz, R.~Pang, W.~Li, and C.~Raffel, ``Monotonic infinite lookback attention for simultaneous machine translation,'' \emph{arXiv preprint arXiv:1906.05218}, 2019.

\bibitem{chiu2017monotonic}
C.-C. Chiu and C.~Raffel, ``Monotonic chunkwise attention,'' \emph{arXiv preprint arXiv:1712.05382}, 2017.

\bibitem{ma2019monotonic}
X.~Ma, J.~Pino, J.~Cross, L.~Puzon, and J.~Gu, ``Monotonic multihead attention,'' \emph{arXiv preprint arXiv:1909.12406}, 2019.

\bibitem{ma2018stacl}
M.~Ma, L.~Huang, H.~Xiong, R.~Zheng, K.~Liu, B.~Zheng, C.~Zhang, Z.~He, H.~Liu, X.~Li \emph{et~al.}, ``Stacl: Simultaneous translation with implicit anticipation and controllable latency using prefix-to-prefix framework,'' \emph{arXiv preprint arXiv:1810.08398}, 2018.

\bibitem{liu2020low}
D.~Liu, G.~Spanakis, and J.~Niehues, ``Low-latency sequence-to-sequence speech recognition and translation by partial hypothesis selection,'' \emph{arXiv preprint arXiv:2005.11185}, 2020.

\bibitem{wang2024simul}
H.~Wang, G.~Hu, G.~Lin, W.-Q. Zhang, and J.~Li, ``Simul-whisper: Attention-guided streaming whisper with truncation detection,'' \emph{arXiv preprint arXiv:2406.10052}, 2024.

\bibitem{li2024transcription}
Y.~Li, X.~Wang, S.~Cao, Y.~Zhang, L.~Ma, and L.~Xie, ``A transcription prompt-based efficient audio large language model for robust speech recognition,'' \emph{arXiv preprint arXiv:2408.09491}, 2024.

\bibitem{dong2020cif}
L.~Dong and B.~Xu, ``Cif: Continuous integrate-and-fire for end-to-end speech recognition,'' in \emph{ICASSP 2020-2020 IEEE International Conference on Acoustics, Speech and Signal Processing (ICASSP)}.\hskip 1em plus 0.5em minus 0.4em\relax IEEE, 2020, pp. 6079--6083.

\bibitem{gao2022paraformer}
Z.~Gao, S.~Zhang, I.~McLoughlin, and Z.~Yan, ``Paraformer: Fast and accurate parallel transformer for non-autoregressive end-to-end speech recognition,'' \emph{arXiv preprint arXiv:2206.08317}, 2022.

\bibitem{ma2024wenetspeech4tts}
L.~Ma, D.~Guo, K.~Song, Y.~Jiang, S.~Wang, L.~Xue, W.~Xu, H.~Zhao, B.~Zhang, and L.~Xie, ``Wenetspeech4tts: A 12,800-hour mandarin tts corpus for large speech generation model benchmark,'' \emph{arXiv preprint arXiv:2406.05763}, 2024.

\bibitem{panayotov2015librispeech}
V.~Panayotov, G.~Chen, D.~Povey, and S.~Khudanpur, ``Librispeech: an asr corpus based on public domain audio books,'' in \emph{2015 IEEE international conference on acoustics, speech and signal processing (ICASSP)}.\hskip 1em plus 0.5em minus 0.4em\relax IEEE, 2015, pp. 5206--5210.

\bibitem{pratap2020mls}
V.~Pratap, Q.~Xu, A.~Sriram, G.~Synnaeve, and R.~Collobert, ``Mls: A large-scale multilingual dataset for speech research,'' \emph{arXiv preprint arXiv:2012.03411}, 2020.

\bibitem{wang2021voxpopuli}
C.~Wang, M.~Riviere, A.~Lee, A.~Wu, C.~Talnikar, D.~Haziza, M.~Williamson, J.~Pino, and E.~Dupoux, ``Voxpopuli: A large-scale multilingual speech corpus for representation learning, semi-supervised learning and interpretation,'' \emph{arXiv preprint arXiv:2101.00390}, 2021.

\bibitem{hu2021lora}
E.~J. Hu, Y.~Shen, P.~Wallis, Z.~Allen-Zhu, Y.~Li, S.~Wang, L.~Wang, and W.~Chen, ``Lora: Low-rank adaptation of large language models,'' \emph{arXiv preprint arXiv:2106.09685}, 2021.

\bibitem{yao2021wenet}
Z.~Yao, D.~Wu, X.~Wang, B.~Zhang, F.~Yu, C.~Yang, Z.~Peng, X.~Chen, L.~Xie, and X.~Lei, ``Wenet: Production oriented streaming and non-streaming end-to-end speech recognition toolkit,'' in \emph{Proc. Interspeech}.\hskip 1em plus 0.5em minus 0.4em\relax Brno, Czech Republic: IEEE, 2021.

\bibitem{yujian2007normalized}
L.~Yujian and L.~Bo, ``A normalized levenshtein distance metric,'' \emph{IEEE transactions on pattern analysis and machine intelligence}, vol.~29, no.~6, pp. 1091--1095, 2007.

\bibitem{cherry2019thinking}
C.~Cherry and G.~Foster, ``Thinking slow about latency evaluation for simultaneous machine translation,'' \emph{arXiv preprint arXiv:1906.00048}, 2019.

\bibitem{chen2024bestow}
Z.~Chen, H.~Huang, O.~Hrinchuk, K.~C. Puvvada, N.~R. Koluguri, P.~{\.Z}elasko, J.~Balam, and B.~Ginsburg, ``Bestow: Efficient and streamable speech language model with the best of two worlds in gpt and t5,'' \emph{arXiv preprint arXiv:2406.19954}, 2024.

\bibitem{yang2024qwen2}
A.~Yang, B.~Yang, B.~Zhang, B.~Hui, B.~Zheng, B.~Yu, C.~Li, D.~Liu, F.~Huang, H.~Wei \emph{et~al.}, ``Qwen2. 5 technical report,'' \emph{arXiv preprint arXiv:2412.15115}, 2024.

\bibitem{jia2024efficient}
J.~Jia, G.~Keren, W.~Zhou, E.~Lakomkin, X.~Zhang, C.~Wu, F.~Seide, J.~Mahadeokar, and O.~Kalinli, ``Efficient streaming llm for speech recognition,'' \emph{arXiv preprint arXiv:2410.03752}, 2024.

\bibitem{tsunoo2024decoder}
E.~Tsunoo, H.~Futami, Y.~Kashiwagi, S.~Arora, and S.~Watanabe, ``Decoder-only architecture for streaming end-to-end speech recognition,'' \emph{arXiv preprint arXiv:2406.16107}, 2024.

\end{thebibliography}

\end{document}